\pdfoutput=1

\documentclass[11pt]{article}

\usepackage[final]{acl}

\usepackage{times}
\usepackage{latexsym}

\usepackage[T1]{fontenc}

\usepackage[utf8]{inputenc}

\usepackage{microtype}

\usepackage{inconsolata}

\usepackage{graphicx}
\usepackage{amsmath} 
\usepackage{amssymb}
\usepackage{booktabs}
\title{EmoFace: Emotion-Content Disentangled Speech-Driven\\ 3D Talking Face Animation}

\author{
 \textbf{Yihong Lin\textsuperscript{1*}},
 \textbf{Liang Peng\textsuperscript{2*}},
 \textbf{Zhaoxin Fan\textsuperscript{3,4}},
 \textbf{Xianjia Wu\textsuperscript{2}},
\\
 \textbf{Jianqiao Hu\textsuperscript{1}},
 \textbf{Xiandong Li\textsuperscript{2$\dagger$}},
 \textbf{Wenxiong Kang\textsuperscript{1$\dagger$}},
 \textbf{Songju Lei\textsuperscript{5}},
\\
\\
 \textsuperscript{1}South China University of Technology,
 \textsuperscript{2}Huawei Cloud,\\
 \textsuperscript{3}Beijing Advanced Innovation Center for Future Blockchain and Privacy Computing, School of \\Artificial Intelligence, Beihang University,\\
 \textsuperscript{4}Hangzhou International Innovation Institute, Beihang University,
 \textsuperscript{5}Nanjing University
\\
 \small{
   \href{mailto:lxdphys@smail.nju.edu.cn}{lxdphys@smail.nju.edu.cn}
   \href{mailto:auwxkang@scut.edu.cn}{auwxkang@scut.edu.cn}
 }
}

\begin{document}
\maketitle
\begin{abstract}
The creation of increasingly vivid 3D talking face has become a hot topic in recent years. Currently, most speech-driven works focus on lip synchronisation but neglect to effectively capture the correlations between emotions and facial motions. To address this problem, we propose a two-stream network called EmoFace, which consists of an emotion branch and a content branch. EmoFace employs a novel Mesh Attention mechanism to analyse and fuse the emotion features and content features. Particularly, a newly designed spatio-temporal graph-based convolution, SpiralConv3D, is used in Mesh Attention to learn potential temporal and spatial feature dependencies between mesh vertices. In addition, to the best of our knowledge, it is the first time to introduce a new self-growing training scheme with intermediate supervision to dynamically adjust the ratio of groundtruth adopted in the 3D face animation task. Comprehensive quantitative and qualitative evaluations on our high-quality 3D emotional facial animation dataset, 3D-RAVDESS ($4.8863\times 10^{-5}$mm for LVE and $0.9509\times 10^{-5}$mm for EVE), together with the public dataset VOCASET ($2.8669\times 10^{-5}$mm for LVE and $0.4664\times 10^{-5}$mm for EVE), demonstrate that our approach achieves state-of-the-art performance.
\end{abstract}

\section{Introduction}

Generating realistic 2D/3D face animations has received great attention in a number of fields, including film production, computer games, virtual reality, education, etc. \cite{ping2013computer,edwards2016jali,wohlgenannt2020virtual}. Recent methods based on deep learning \cite{karras2017audio,cudeiro2019capture,richard2021meshtalk,fan2022faceformer,xing2023codetalker,peng2023selftalk,wu2023speech,stan2023facediffuser} produce impressive 3D face animations with significant savings in time and labour compared to manual production, making them preferred for academic research and commercial exploration. However, there are still some situations remained to be adequately addressed, for example, the relationship between emotions and facial expressions. Most of the preceding speech-driven methods \cite{cudeiro2019capture,richard2021meshtalk,fan2022faceformer,xing2023codetalker,peng2023selftalk} focus more on achieving high-quality lip synchronisation, while ignoring the impact of emotion on facial animations. Recognizing the significance of emotion, EmoTalk \cite{peng2023emotalk} and EMOTE \cite{danvevcek2023emotional} disentangle emotional and content information in speech to generate high-quality emotional 3D facial animations. Nevertheless, EmoTalk only uses emotion as the driving source, regardless of content, thereby limiting its performance. EMOTE doesn't achieve end-to-end training and adopts a FLAME \cite{li2017learning} parameter-based method that theoretically does not control motion as accurately as vertex-based methods because the abstract coefficient estimation ignores spatial correlations.

In this paper, we propose a novel vertex-based autoregressive emotional speech-driven 3D face animation approach. We first disentangle emotion and content from speech and extract their features separately, then a latent space decoder is used to obtain the emotion-based and content-based vertex offsets, which are eventually integrated by Mesh Attention to predict the final offsets. Due to that the fusion weights for the emotion branch and the content branch are variable in both temporal and spatial domains, we add a 3D graph-based convolution operator SpiralConv3D in Mesh Attention for effective feature exaction. Moreover, during the training process, it is observed that the autoregression scheme of the transformer decoder results in error accumulation, which ultimately leads to a deterioration in the output and increases the training time. Although, existing teacher-forcing scheme can help to alleviate this problem, exposure bias may occur when switching from training to inference due to inability to access real historical data. Inspired by scheduled sampling \cite{scheduled-transformers-2019}, we propose a self-growing scheme that gradually adjusts the ratio of groundtruth provided during training. In order to better train our emotion-content disentanglement model, we construct a high-quality dataset 3D-RAVDESS by reconstructing reliable 3D faces from 2D dataset RAVDESS \cite{livingstone2018ryerson}. Extensive qualitative and quantitative experiments demonstrate that our method outperforms current state-of-the-art methods for better generation of facial expressions and lip synchronisation. In summary, our main contributions are as follows:
\begin{itemize}

\item We provide a new high-quality emotional speech-driven dataset, 3D-RAVDESS.

\item We design a two-stream network, EmoFace, to obtain the emotion-based and content-based vertex offsets, respectively, which are dynamically fused by a novel Mesh Attention with SpiralConv3D that efficiently extracts features in the spatio-temporal domain.

\item We introduce intermediate supervision to guide two branches respectively and propose a self-growing training scheme to gradually increase the difficulty of prediction, thus improving the robustness of the model.

\item Extensive experiments demonstrate the superiority of EmoFace over existing SOTA methods in terms of full face realism, emotion expression and lip synchronisation on both 3D-RAVDESS and VOCASET.
\end{itemize}

\section{Related Work}

\subsection{Speech-Driven 3D Talking Face}

Speech-driven 3D talking face generation is a task to generate realistic facial animations based on speech \cite{karras2017audio,lahiri2021lipsync3d,pham2017speech,taylor2017deep}. A number of rule-based approaches have been employed to generate 3D motions, including the capture of the relationship between visemes \cite{mattheyses2015audiovisual} and facial action units (FAUs) \cite{ekman1978facial}, as well as the establishment of a phoneme-viseme mapping that has yielded highly promising results in early studies \cite{edwards2016jali,xu2013practical}. Nevertheless, these methods do not easily transfer to new faces as they heavily rely on manual crafting and consume an excessive amount of time.

In contrast to rule-based methods, deep learning-based 3D face animation methods resort to a data-driven framework. Most of them take both speech and a static 3D mesh template as input to generate realistic 3D face animations. In an early work, VOCA \cite{cudeiro2019capture} proposes a temporal convolutional neural network with an open-source 4D dataset VOCASET, \cite{cudeiro2019capture}, which has become a valuable resource for subsequent studies. Building on this foundation, MeshTalk \cite{richard2021meshtalk}, FaceFormer \cite{fan2022faceformer}, and CodeTalker \cite{xing2023codetalker} incorporate motion priors to improve animation quality. MeshTalk notes the importance of facial motion in audio-uncorrelated regions and adopts categorical latent space to learn discrete motion priors. FaceFormer focuses on the long temporal sequences and successfully uses a transformer decoder \cite{vaswani2017attention} to obtain contextual information to generate sequential mesh sequences. CodeTalker introduces VQ-VAE \cite{van2017neural} to learn discrete motion priors for smooth facial motion generation. More recently, SelfTalk \cite{peng2023selftalk} proposes consistency loss to improve the generation quality. Other works \cite{stan2023facediffuser,chen2023diffusiontalker,sun2024diffposetalk} introduce diffusion model into 3D facial animation to enhance motion diversity.

However, all these approaches have mainly focused on lip synchronisation but have not paid attention to the correlation between emotions and facial expressions. To address this problem, EmoTalk \cite{peng2023emotalk} and EMOTE \cite{danvevcek2023emotional} both propose speech-driven content-emotion disentanglement pipelines. Nevertheless, EmoTalk does not utilize content information and only focuses on the driving effect of emotional information, while the effectiveness and efficiency of EMOTE are limited by FLAME parameters and two-stage training pipelines, respectively.

\begin{figure*}[t]
    \centering
    \includegraphics[scale=0.58]{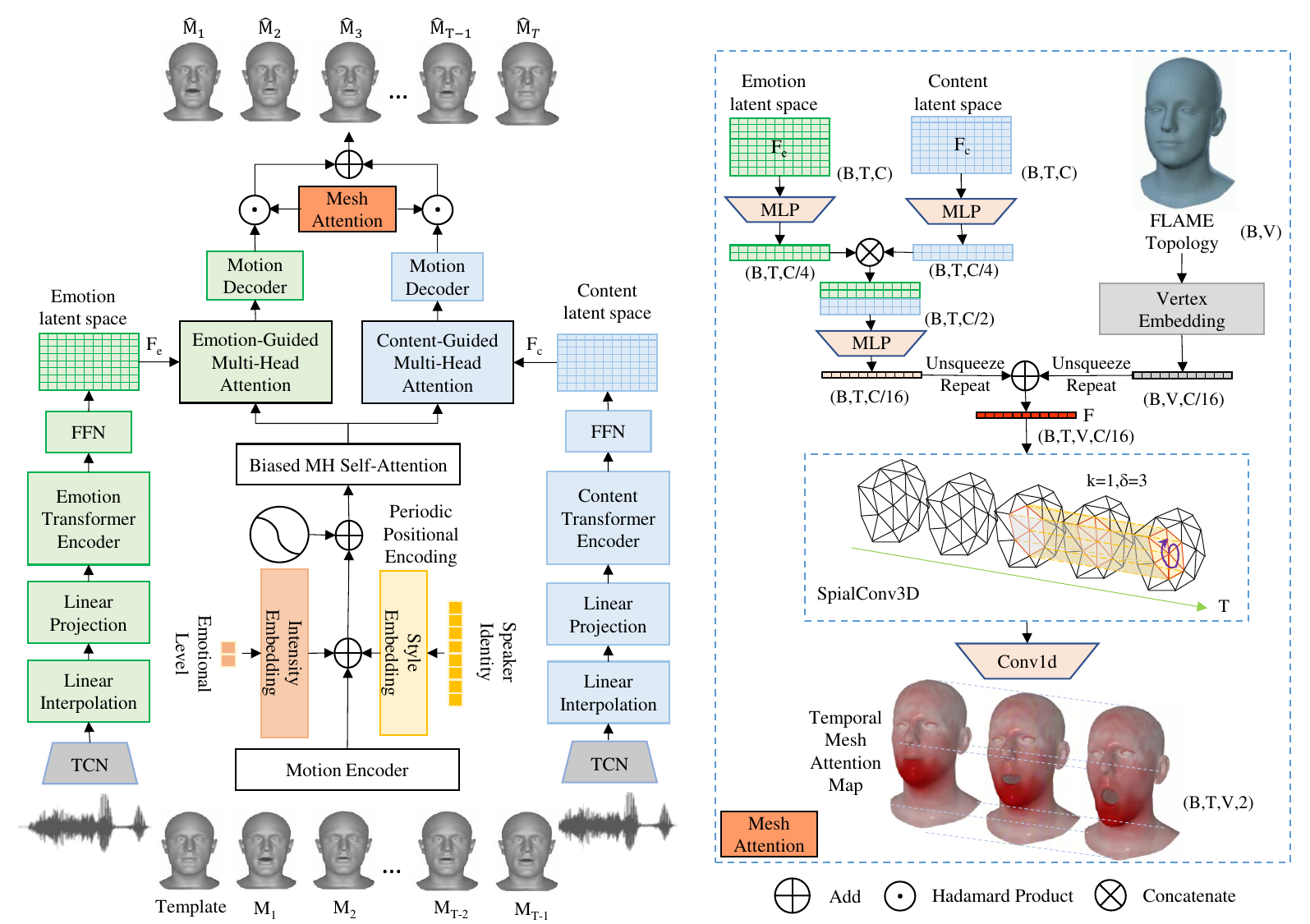}
    \caption{The overall framework of EmoFace. The emotion and content branches disentangle the information in speech, while Mesh Attention fuses these two branches to obtain the final result. The entire framework is end-to-end, thus allowing for efficient training and inference. $\hat M_i$ denotes the $i$-th predicted motion and $M_i$ denotes the $i$-th reference motion. $k$ and $\delta$ represent the spatial and the temporal neighbourhood of SpiralConv3D respectively.}
    \vspace{-0.5cm}
    \label{fig1}
\end{figure*}

\subsection{Spiral Convolution}

Defferrard et al. \cite{defferrard2016convolutional} present a graph convolutional network (GCN) based on spectral filtering for non-Euclidean graph structure with the same linear complexity as the classical CNN. Kolotouros et al. \cite{kulon2020weakly} devise a simple model consisting of an encoder and a decoder with spiral convolution added to the decoder, which can directly learn the mapping relationships from 2D images to 3D meshes. Since the graph convolution operator is intuitive enough for dealing with 3D mesh vertices in the spatial domain, Masci et al. \cite{masci2015geodesic} advance the idea of spatial graph convolution with sampling of the graph signal. Lim et al. \cite{lim2018simple} propose spiral convolution based on graph convolution to handle 3D mesh vertices in the spatial domain. Gong et al. \cite{gong2019spiralnet++} introduce SpiralNet++, which utilizes truncated spiral lines to constrain the number of sampled vertices, while incorporating hollow spiral lines to enhance the receptive field. Inspired by these previous works, we use spiral convolution in a 3D facial animation generation task to allow EmoFace to better learn the spatial as well as temporal associations of mesh vertices.

\section{Methods}

\subsection{Overview}

The overall pipeline is shown in Figure \ref{fig1}. In order to generate vivid emotional 3D talking faces, we propose EmoFace, a model that can sufficiently learn emotions from speech and generate talking face with rich expressions. The input of EmoFace consists of the speech sequence $A_{1:T}=(a_{1},a_{2},{\ldots},a_{T})$, the emotional level $l \in \mathbb{R}^{2}$, the speaking style $s \in \mathbb{R}^{n}$ and the character template $Y_c \in \mathbb{R}^{V \times 3}$, where $n$ is the number of speaking styles, $T$ is the length of the mesh sequence, and $V$ is the number of mesh vertices. The emotion level and the speaking style are encoded as one-hot vectors. Inspired by \cite{ji2021audio} and \cite{peng2023emotalk}, we use a similar emotion-content disentanglement module that uses pre-trained speech feature extractors wav2vec2.0 \cite{baevski2020wav2vec} to reduce the difficulty of learning the mapping between speech and emotional facial expressions. Specifically, the emotion branch and the content branch firstly disentangle emotion-related features $F_e$ and content-related features $F_c$ from the speech sequence $A_{1:T}$. Then, emotion-related features $F_e$ and content-related features $F_c$ drive the given template vertices to generate offsets $\Delta M^{e}_{1: T} = (\Delta m^{e}_{1}, \Delta m^{e}_{2},{\ldots}, \Delta m^{e}_{T})$ and $\Delta M^{c}_{1: T} = (\Delta m^{c}_{1}, \Delta m^{c}_{2},{\ldots}, \Delta m^{c}_{T})$, respectively, using the same transformer decoder as FaceFormer, which contains a multi-head cross-attention module to align the audio and motion modalities as well as a multi-head self-attention module to learn the dependencies between each frame of the past facial motions. Finally, Mesh Attention is provided to fuse the predictions of the two branches in temporal and spatial domains to obtain the final prediction $\Delta M_{1: T}$.

\subsection{Mesh Attention}
In order to fuse the prediction results of the emotion branch and content branch, we propose a novel spatio-temporal attention module Mesh Attention that analyses the fusion weight. Different from conventional MLP-based motion decoding methods that suffer from mesh Non-Euclidean topology destruction and explicit geometric constraint absence, Mesh Attention employs the core operator SpiralConv3D to directly model spatial vertex connectivity within frames and temporal vertex correspondence across frames. This approach performs spatiotemporal aggregation of neighboring vertices' geometric and semantic features while maintaining the mesh's local manifold structure and physical constraints (e.g., smoothness, symmetry), ultimately generating a Temporal Mesh Attention Map that quantifies the impact of two branches on final predictions. 

Firstly, $F_{e1} \in \mathbb{R}^{B \times T \times C} $ and $F_{c1} \in \mathbb{R}^{B \times T \times C}$ go through one layer of MLP to get $F_{e1} \in \mathbb{R}^{B \times T \times C/4} $ and $F_{c1} \in \mathbb{R}^{B \times T \times C/4}$ with lower channel dimensions, and then they are concatenated in channel dimensions, followed by another layer of MLP to further reduce the channel number to get the fusion feature $F_{audio} \in \mathbb{R}^{B \times T \times C/16}$. Secondly, we perform the vertex position embedding of the FLAME topology to obtain the position embedding feature $F_{vertices-emb} \in \mathbb{R}^{B \times V \times C/16}$. Furthermore, to obtain the temporal features of the mesh vertices, we integrate the fusion feature $F_{audio}$ and the position embedding feature $F_{vertices-emb}$. Specifically, the fusion feature $F_{audio}$ is unsqueezed in the second dimension and then repeated $V$ times to obtain $F^{\prime}_{audio} \in \mathbb{R}^{B \times V \times T \times C/16}$. Similarly, $F_{vertices-emb}$ is unsqueezed in the first dimension and then repeated $T$ times to obtain $F^{\prime}_{vertices-emb} \in \mathbb{R}^{B \times V \times T \times C/16}$. The obtained features are summed eventually to get the integrated temporal features of the vertices $F \in \mathbb{R}^{B \times V \times T \times C/16}$. 

\begin{figure*}[t]
    \centering
    \includegraphics[scale=0.58]{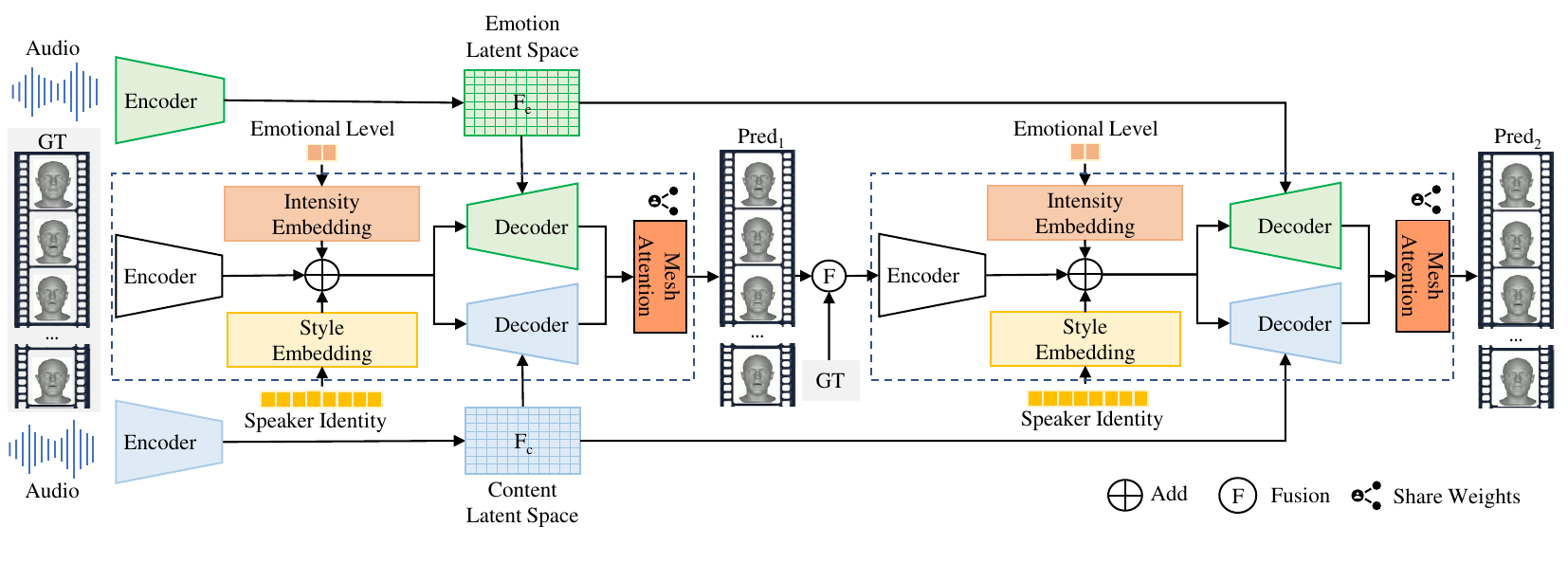}
    \caption{Self-growing scheme. In the first stage, the model inputs all groundtruth frames at once and directly predicts the next frame corresponding to each input frame. In the second stage, the input is changed to a fusion of the groundtruth frames and the predicted frames from the previous stage, and the same prediction process is applied to obtain the final prediction results.}
    \label{fig2}
    \vspace{-0.5cm}
\end{figure*}

In purpose of further exploring the temporal-spatial connection between vertices, we design a new graph-based convolution operator, 3D spiral convolution (SpiralConv3D). In spatial domain, it determines the convolution centre and generates a series of enumerated vertices based on adjacency, followed by 1-ring vertices, 2-ring vertices, and so on, until all vertices containing k rings are included. SpiralConv3D determines the adjacency as follows:
\begin{align}\label{eq1}
    0{-ring}(v)&={v}, \nonumber \\
    (k+1){-}ring(v)&=\mathcal{N}(k{-ring}(v)) \setminus k{-disk}(v), \nonumber \\
    k{-disk}(v)&=\cup_{i=0,...,k}i{-ring}(v), 
\end{align}
where $\mathcal{N}(V)$ selects all vertices in the neighborhood of any vertex in set $V$. In temporal domain, SpiralConv3D considers how the vertices have changed between the past $T$ frames, and creates connections to the corresponding vertices between different frames:
\begin{equation}\label{eq2}
     cnt_{\delta}(V_{ti})=
    \left\{\begin{aligned}
    \{ V_{t-\delta+1,i},\ldots,V_{t-1,i},V_{t,i}) \}& , t \geq \delta\\
    \{ Pad(\delta-t),V_{1,i},\ldots,V_{t,i}\}& , t < \delta
    \end{aligned}\right. ,
\end{equation}
where $V_{ti}$ is the $i$-th vertex of moment $t$, $0 \leq t \leq T$ and $0 \leq i \leq 5023$. Meanwhile, $cnt_{\delta}(V_{ti})$ denotes the connection between $V_{ti}$ and the vertices of past $\delta$ frames, and padding is used when $t < \delta$. Similar to 3D ConvNets \cite{tran2015learning}, SpiralConv3D considers both temporal and spatial correlations. In contrast, SpiralConv3D applies this spatio-temporal modeling to mesh sequences with a sparse spatial distribution and a dense temporal distribution, considering convolution as the use of fully connected layers for feature fusion:
\begin{align} \label{eq3}
    S\!piralConv3D(v) = &W(f(cnt_{\delta}(k{-disk}(v))))\nonumber\\&+b,
\end{align}
where $f$ denotes the feature extractor, $W$ and $b$ are learnable weights and bias. To our knowledge, this work exploits 3D GraphConv in the context of supervised training datasets and modern deep architectures to achieve the best performance on 3D facial animation.

\subsection{Training and Testing}

\subsubsection{Self-growing Scheme}

During the training phase, we adopt a novel self-growing scheme instead of teacher-forcing or autoregression scheme, as shown in Figure \ref{fig2}. Self-growing scheme divides the training process in each epoch into two stages. In the first stage, the model inputs all groundtruth frames at once, and then uses Temporal Bias \cite{fan2022faceformer} to eliminate the influence of future frames and directly predicts the next frame corresponding to each input frame. In the second stage, the model adopts the fusion strategy to obtain new input data and repeats the prediction operation in the previous stage. The fusion strategy is adapted according to the training process. Specifically, in the first few $\theta$ epochs, we keep all the groundtruth frames $GT$ as inputs; in the remaining epochs, the first half gradually replace the groundtruth frames $GT$ with the predicted frames $P$ from the previous stage using the cosine function, and the second half directly abandon the groundtruth frames. The fusion strategy can be denoted as follows:

\begin{align}\label{4}
    I_{n,t}=\left\{
                \begin{array}{ll}
                  GT_t, n<\theta \text{\enspace} or \text{\enspace} rand() < cos(\frac{\pi(n-\theta)}{N-\theta})\\
                  P_t, n>\frac{(N+\theta)}{2} \text{\enspace} or \text{\enspace} rand() \geq  cos(\frac{\pi(n-\theta)}{N-\theta})
                \end{array},
              \right.
\end{align}
where $n$ and $t$ denote the $n$-th epoch and the $t$-th frame, respectively. $N$ means the total number of epochs and $rand()$ generates a random number from a uniform distribution over the interval [0, 1]. Self-growing scheme fully guides in the initial epochs and then gradually reduces the guidance and accepts its own outputs, which weakens the impact of error accumulation and improves the robustness and generation quality. 

During inference, EmoFace autoregressively predicts the mesh sequence corresponding to previous 3D talking faces. Specifically, at each moment $t$, EmoFace predicts the face motion $M_t$ conditioned on the raw audio $A$, the prior sequence of face motions $M_{\leq t}$ , the speaking style $s_n$, and the emotion level $l_n$ at each moment. The $s_n$ and $l_n$ are determined by the speaker, and thus altering the one-hot identity and intensity vectors can manipulate the output in different styles and emotional levels.

\subsubsection{Training Strategy of Emotion-Content Disentanglement Module} 

We set a pseudo-training pair of samples with the same content but different emotions. The emotion features and content features are extracted respectively and the two emotion features are exchanged as input of transformer decoder to implement cross reconstruction. Since that we expect both the emotion branch and the content branch to be as strong as possible in modeling the speech-mesh mapping relationship, our emotion branch and content branch make separate predictions for the mesh offsets of the next frame. We incorporate the intermediate supervision into the predictions of two separate branches. However, both branches have significant limitations in their respective abilities to model facial motions. On the one hand, although the features extracted from the content branch are strongly correlated with the lips, the inability to extract long-term emotional features and the lack of the type and intensity of emotion are not conducive to predicting the magnitude of mouth opening. On the other hand, though the features extracted from the emotion branch are strongly correlated with the motions of the eyes and the face expressions, the lack of short-term content features results in insufficiently accurate predictions of the lips. Thus, as mentioned before, Mesh Attention is designed to fuse the driving results of the two branches in temporal and spatial domains to obtain the final prediction. Our self-reconstruction loss, cross-reconstruction loss, velocity loss and classification loss are as follows:

\begin{align}\label{6}
{L}_{self} &= \left \| M^{e}_{c1,e1}- \hat{M}_{c1,e1}\right \|_1 +\left \| M^{c}_{c1,e1}- \hat{M}_{c1,e1}\right \|_1\nonumber\\& + \left \| M_{c1,e1} - \hat{M}_{c1,e1}\right \|_1, \\
{L}_{cross} &= \left \| M^{e}_{c1,e2} - \hat{M}_{c1,e2}\right \|_1+\left \| M^{c}_{c1,e2} - \hat{M}_{c1,e2}\right \|_1\nonumber\\
&+\left \| M_{c1,e2} - \hat{M}_{c1,e2}\right \|_1, \\
{L}_{vel} &= \left \| M^{t}_{c1,e1}-M^{t-1}_{c1,e1},\hat{M}^{t}_{c1,e1}-\hat{M}^{t-1}_{c1,e1}\right \|_1\nonumber\\
&+\left \| M^{t}_{c1,e2}-M^{t-1}_{c1,e2},\hat{M}^{t}_{c1,e2}-\hat{M}^{t-1}_{c1,e2}\right \|_1, \\
{L}_{cls} &= - \sum_{i} \sum_{\phi=1}^{N_e}\left ( y_{i\phi} * \log p_{i\phi} \right ),
\end{align}
where $M_{cx,ey}$ represents predicted mesh sequence with content $x$ and emotion $y$. $M^c$ and $M^e$ represents predicted mesh sequence of content branch and emotion branch, respectively. $M^t$ means the $t$-th frame of mesh sequence, $\hat{M}$ means the groundtruth, $N_e$ represents the number of distinct emotion categories, $y_{i\phi}$ is the observation function that determines whether sample $i$ carries the emotion label $\phi$, and $p_{i\phi}$ denotes the predicted probability that sample $i$ belongs to class $\phi$. Note that the first two terms of self-reconstruction loss and cross-reconstruction loss are intermediate supervision of the two branches. The overall function is given by:

\begin{align}\label{7}
L =\lambda_{1}L_{self} + \lambda_{2}L_{cross} +\lambda_{3}L_{vel}+\lambda_{4}L_{cls}, 
\end{align}
where $\lambda_{1}=\lambda_{2}=1000$, $\lambda_{3} = 500$ and $\lambda_{4} = 0.0001$ in all of our experiments.

\section{Experiments}

\subsection{Experimental Settings}
We employ two datasets, including the non-emotional dataset VOCASET and the emotional dataset 3D-RAVDESS, where the 3D-RAVDESS dataset is the dataset we constructed.

\indent\textbf{VOCASET dataset}. VOCASET \cite{cudeiro2019capture} consists of 480 face mesh sequences from 12 subjects. Each mesh sequence is 60fps and is between 3 and 4 seconds long in duration. Meanwhile, each 3D face mesh has 5023 vertices. We follow the data configuration of VOCA for a fair comparison.

\indent\textbf{3D-RAVDESS Dataset}. The RAVDESS \cite{livingstone2018ryerson} is a multimodal emotion recognition dataset. As shown in Table \ref{table1}, the dataset contains 24 actors (12 male and 12 female), each actor has 60 sentences with a total of 1440 face mesh sequences and corresponding speech. Each actor in this dataset provides data on different emotion categories and emotion levels, including neutral, calm, happy, sad, angry, fearful, disgusted, and surprised. In this case, the data corresponding to the first 20 subjects of the dataset are used for training, the 21st and 22nd subjects are used for validation, and the last two subjects are used for testing. We first process 1440 videos from the original RAVDESS dataset to convert the frame rate to 30 fps. Then, the emotional 3D faces are reconstructed using EMOCA \cite{danvevcek2022emoca} to obtain a sequence of 5023 mesh vertices under the FLAME network topology. Due to the jittery results, we use Kalman Filter \cite{kalman1960new} on the FLAME parameters and fix the last three of the pose parameters to obtain 3D head mesh sequences with smooth front view. Our 3D-RAVDESS dataset consists of these mesh sequences and the corresponding speech from the original 2D dataset.

\begin{table*}[t]
    \centering
    \scalebox{0.8}{
        \begin{tabular}{@{}cccccccc@{}}
        \toprule
Subject  & Gender Ratio & Text & Emotion & Sentence & FPS & Topology & Training:Validation:Testing   \\
\midrule
24  & 1:1  & 60 & 8 & 1440 & 30 & FLAME & 10:1:1 \\
        \bottomrule
        \end{tabular}
        }
    \caption{Statistics of 3D-RAVDESS dataset.}
    \label{table1}
\end{table*}

\begin{figure*}[t]
    \centering
    \includegraphics[scale=0.5]{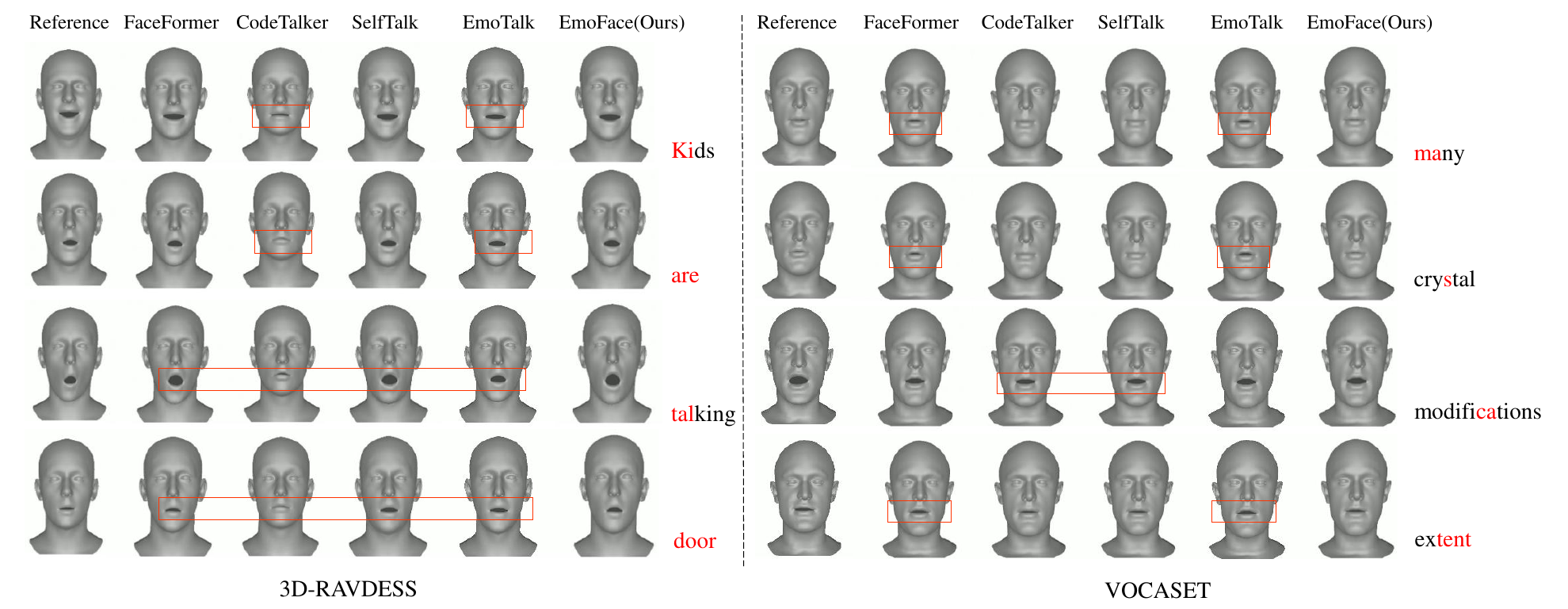}
    \caption{Qualitative comparison of the facial movements of the different methods on 3D-RAVDESS (left) and VOCASET (right). On 3D-RAVDESS, we generate facial animations of saying the sentence “Kids are talking by the door.” with surprised. On VOCA-Test, facial animations of saying the sentence “How many crystal modifications of uranium hydride are extent?” without emotion are generated. Significant differences in the lip region are denoted by red boxes. EmoFace generates more realistic facial movements that match the speech, whether it's emotional or not.}
    \label{fig3}
\end{figure*}

\subsection{Quantitative evaluation}
To evaluate the lip synchronisation, we compute the lip vertex error (LVE) that is used in previous work \cite{fan2022faceformer}. This metric computes the maximum $\ell_{2}$ error among all lip vertices in the test set and averages $\ell_{2}$ error across all frames. Additionally, the emotional vertex error (EVE) \cite{peng2023emotalk} is used to reflect the full emotional expression. It measures the maximum $\ell_{2}$ error of all eye and forehead vertices in the test set and averages $\ell_{2}$ error of them. Table \ref{table2} demonstrates significant advantages of our algorithm in handling emotional speech-driven 3D face. In particular, our LVE and EVE on 3D-RAVDESS dataset is 20\% and 35\% lower than SelfTalk, respectively.

\begin{table}[t]
    \centering
        \scalebox{0.8}{
        \begin{tabular}{@{}lcccc@{}}
        \toprule
         & \multicolumn{2}{c}{VOCASET} &  \multicolumn{2}{c}{3D-RAVDESS} \\
         \midrule
Methods       & LVE $\downarrow$     & EVE $\downarrow$  & LVE $\downarrow$     & EVE $\downarrow$  \\
\midrule
VOCA \shortcite{cudeiro2019capture}         & 4.9245    & -- & --      & -- \\
MeshTalk \shortcite{richard2021meshtalk}     & 4.5441    & -- & --      & -- \\
FaceFormer \shortcite{fan2022faceformer}   & 4.1090    & 0.4858 & 5.8462      & 1.4149 \\
CodeTalker \shortcite{xing2023codetalker}   & 3.9445    & 0.5074 & 15.6160      & 3.4675 \\
EmoTalk \shortcite{peng2023emotalk}      & 3.7798    & 0.4862 & 6.6076       & 1.5994 \\
SelfTalk \shortcite{peng2023selftalk}     & 3.2238    & 0.4562 & 6.2560       & 1.5685 \\
TalkingStyle \shortcite{talkingstyle}   & 3.5245    & 0.4686 &  5.9737     & 1.4471 \\
\textbf{EmoFace(Ours)} & \textbf{2.8669}    & \textbf{0.4664} & \textbf{4.8863} & \textbf{0.9509} \\
        \bottomrule
        \end{tabular}
}     
    \caption{Quantitative evaluation results on VOCASET and 3D-RAVDESS.(For better visualization, we scale up the LVE and EVE by a factor of $10^{-5}$.)}    
    \label{table2}
\end{table}

\subsection{Qualitative evaluation}
We qualitatively compare the driving effects of cross-identity between different models by driving the new 3D character templates in the test set with the speaking style of the training characters. The left side of Figure \ref{fig3} shows the driving results of different models for the same speech with strong surprised emotion, and the right side shows the driving results of different models in non-emotional speech. Since that the speaking style of the new characters is unknown in the test data, the model may be driven slightly differently to the real results. In terms of lip synchronisation, EmoFace shows a greater amplitude of movement, which is particularly evident in the lip shapes for “are", “talking" and “door". This provides a more accurate reflection of surprise and is more consistent with real movements of the lips. In addition, it also has a higher degree of mouth closure than other methods, for example, when pronouncing “many", “crystal" and “extent". Furthermore, the full face comparison in Figure \ref{fig3} indicates that EmoFace drives more obvious and natural expressions. To further demonstrate the superiority of our algorithm for both emotional and non-emotional audio inputs, a supplementary video is provided for more detailed comparisons.

\subsection{Ablation experiments}
In this section, we conduct ablation experiments to investigate the impact of our training strategy and Mesh Attention. All ablation experiments are conducted on the 3D-RAVDESS dataset.

\begin{table}[t]

\centering

\scalebox{0.8}{
\begin{tabular}{lcc}
\toprule
Modules & \begin{tabular}[c]{@{}c@{}}LVE  \\ ($\times 10^{-5}$mm)\end{tabular}   & \begin{tabular}[c]{@{}c@{}}EVE  \\ ($\times 10^{-5}$mm)\end{tabular} \\
\hline
w/o self-growing & 5.3898   & 1.6794 \\
w/o Mesh Attention & 10.7420   & 1.0965 \\
\textbf{Ours} & \textbf{4.8863} & \textbf{0.9509} \\
\hline
\end{tabular}
}
\caption{Ablation study for our components. We show the LVE and EVE in different cases.}
\label{table3}
\end{table}

\subsubsection{Impact of self-growing scheme.}
We train our method using teacher-forcing without self-growing and obtain higher LVE and EVE, as shown in Table \ref{table3}. We believe the reason is that guiding is too strong, leading to poor robustness, and the prediction errors in the previous frames accumulate and affect the subsequent frames. However, with a gradually weakened guiding in the self-growing scheme, EmoFace is trained to take the errors of the previous frames into account, which is similar to the situation during inference.

\subsubsection{Impact of Mesh Attention.}
We compare the difference in effectiveness between the Mesh Attention module and the Add operation. Table \ref{table3} also demonstrates that directly adding up the gains predicted by the two branches without Mesh Attention leads to a great degradation of model performance, whereas a weighted sum seems more reasonable. This indicates that the contribution of the two branches differs in the different head regions.

\subsubsection{Hyperparameters of SpiralConv3D.}
A series of ablation experiments on dilation, kernel size and time duration are provided, as shown in Table \ref{table4}. In terms of dilation coefficients, the third to fifth rows show that proper dilation can improve performance, while excessive dilation leads to performance degradation. SpiralConv3D may under-consider nearby vertex features with too large dilation coefficient. Furthermore, with respect to kernel size, the first to third rows indicate that large convolution kernels bring high generation quality. SpiralConv3D with a larger kernel size integrates vertex features within a larger spatial receptive field, which has a significant effect on the performance improvement. Finally, the fourth and the last two rows demonstrate the importance of large temporal receptive field. A larger time duration means that the connectivity relationship of the vertices between more video frames are considered. Time duration coefficient of 1 degrades SpiralConv3D to SpiralConv2D, which only spatially fuses the features and performs worse than configuration of the fourth row. Based on the results of the quantitative comparison, we adopt the configuration of the fourth row.

\begin{table}[t]

\centering

\begin{center}
\scalebox{0.8}{
\begin{tabular}{ccccc}
\toprule
Dilation & \begin{tabular}[c]{@{}c@{}}Kernel  \\ Size\end{tabular} & \begin{tabular}[c]{@{}c@{}}Time  \\ Duration\end{tabular} & \begin{tabular}[c]{@{}c@{}}LVE  \\ ($\times 10^{-5}$mm)\end{tabular}   & \begin{tabular}[c]{@{}c@{}}EVE  \\ ($\times 10^{-5}$mm)\end{tabular} \\
\hline
1 & 5 & 3 & 5.3472 & 1.1495 \\
1 & 9 & 3 & 5.2177 & 1.1089 \\
1 & 18 & 3 & 4.7527 & 1.0244 \\
\textbf{2} & \textbf{18} & \textbf{3} & \textbf{4.8863} & \textbf{0.9509} \\
4 & 18 & 3 & 5.1323 & 1.1390 \\
2 & 18 & 1 & 5.3476 & 1.1715 \\
2 & 18 & 2 & 5.5372 & 1.2144 \\
\hline
\end{tabular}
}
\end{center}
\caption{Ablation study of SpiralConv3D.}
\label{table4}
\end{table}

\begin{figure}[t]
    \centering
    \includegraphics[scale=0.18]{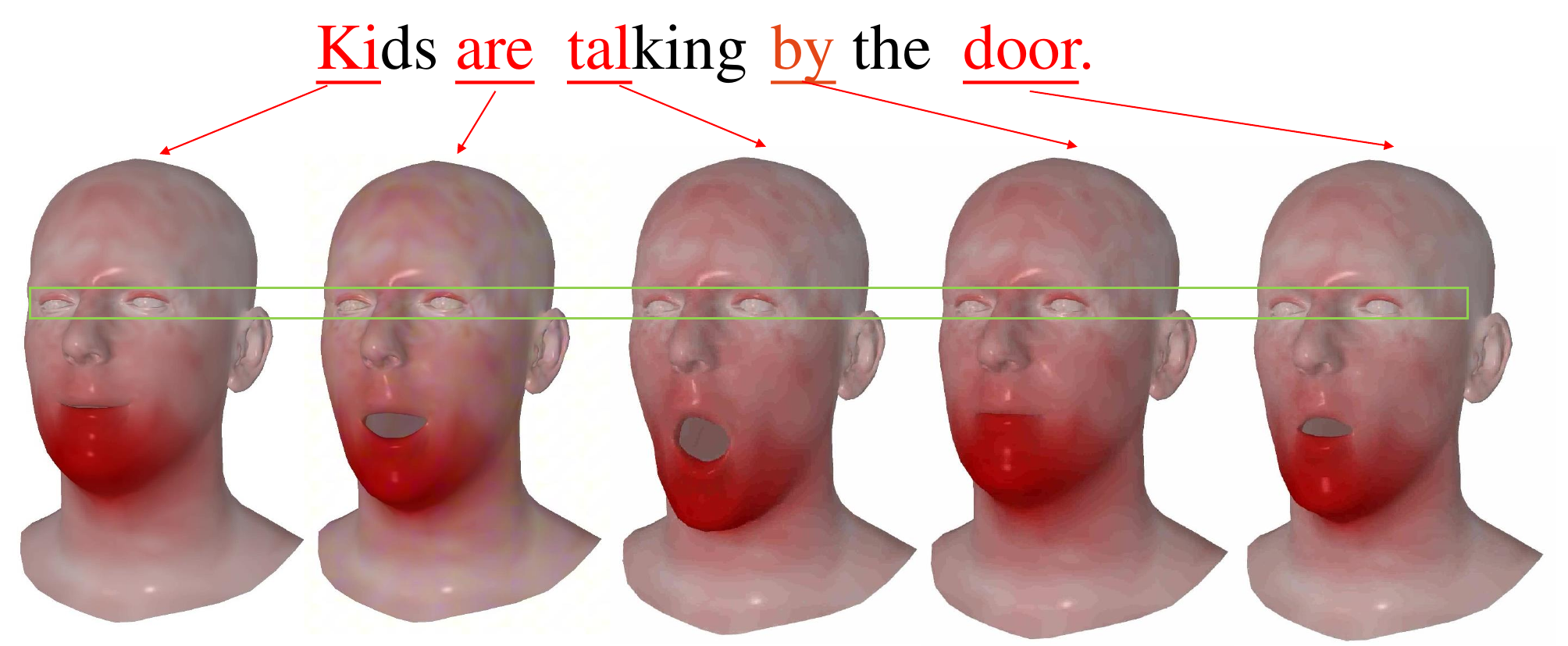}
    \caption{Visualization of the importance of emotional information for facial regions. The eyes, mouth and jaw are strongly correlated with emotions.}
    \vspace{-0.5cm}
    \label{fig4}
\end{figure}

\subsection{Visualization}
We believe that although the lips are directly controlled by the content of the speech, the gain from emotion is also quite important. Hence, we visualize the attention weight of the emotion branch with darker color representing larger weights, as shown in Figure \ref{fig4}. It is obvious that emotion brings great effects on the motion amplitude and direction in the mouth, jaw and eye regions. In other words, these regions are strongly correlative to emotion.

The Supplementary Video shows the animation of the content branch. It performs normally in non-silent frames but shows obvious lip jitter in silent frames, which also leads to unstable Addition output. However, EmoFace dynamically combines two branches with Mesh Attention to obtain better animation results in terms of lip synchronization, expression synchronization, movement amplitude and coherence. More results under VOCASET-Test, 3D-RAVDESS-Test and long speech in the wild, performance on different emotions and comparative demonstrations of ablation experiments are shown in the Supplementary Video.

\section{Conclusion}

In this paper, we present EmoFace, an emotional speech-driven 3D talking face model. Our model disentangles the emotion and content from the speech and predicts the mesh offsets driven by emotion and content, respectively. To further improve the prediction accuracy, new Mesh Attention integrates the output mesh offsets. Particularly, a novel graph-based SpiralConv3D is adopted to fuse spatio-temporal features of mesh sequences. Extensive experiments conducted under self-growing scheme using both public dataset VOCASET and newly constructed high-quality dataset 3D-RAVDESS demonstrate that our model outperforms existing state-of-the-art methods and receives better user experience feedback.

\section*{Limitations}

Although our model gets state-of-the-art results, there are still some limitations to address in future work. On the one hand, speech driven methods do not model expressions and motions that are unrelative to audio, for example, eye blinks. Our next work will take the video input into account to obtain better animation results. On the other hand, the Mesh Attention proposed in our work meets large amount of computation because of the 3D spiral convolution operator. More explorations will be done to reduce the computation costs.

\section*{Ethical Statement}

The existing scientific artifacts used in this work are consistent with their intended use and our work is for research purposes only and should not be used outside of research contexts. In addition, since face data can be used for generating content that may jeopardize privacy, we must act responsibly by considering the aspects related to privacy and ethics.

\bibliography{custom}

\appendix

\section{Appendix}
\label{sec:appendix}


\subsection{Training Details}
Our method takes speech data as input and also provides mesh templates, emotion levels, and speaker IDs as conditions. The sampling rate of the speech is $16kHz$ and the frame rate of the mesh sequence is 30 frames per second. During training, the model is end-to-end optimised using the Adam optimizer. The learning rate and batch size are set to $10^{-4}$ and $1$ respectively. The model is trained on a single NVIDIA A100 with 200 epochs on VOCASET and 3D-RAVDESS.

\subsection{Emotion-Content Disentanglement Module}
\begin{figure*}[t]
    \centering
    \includegraphics[scale=0.4]{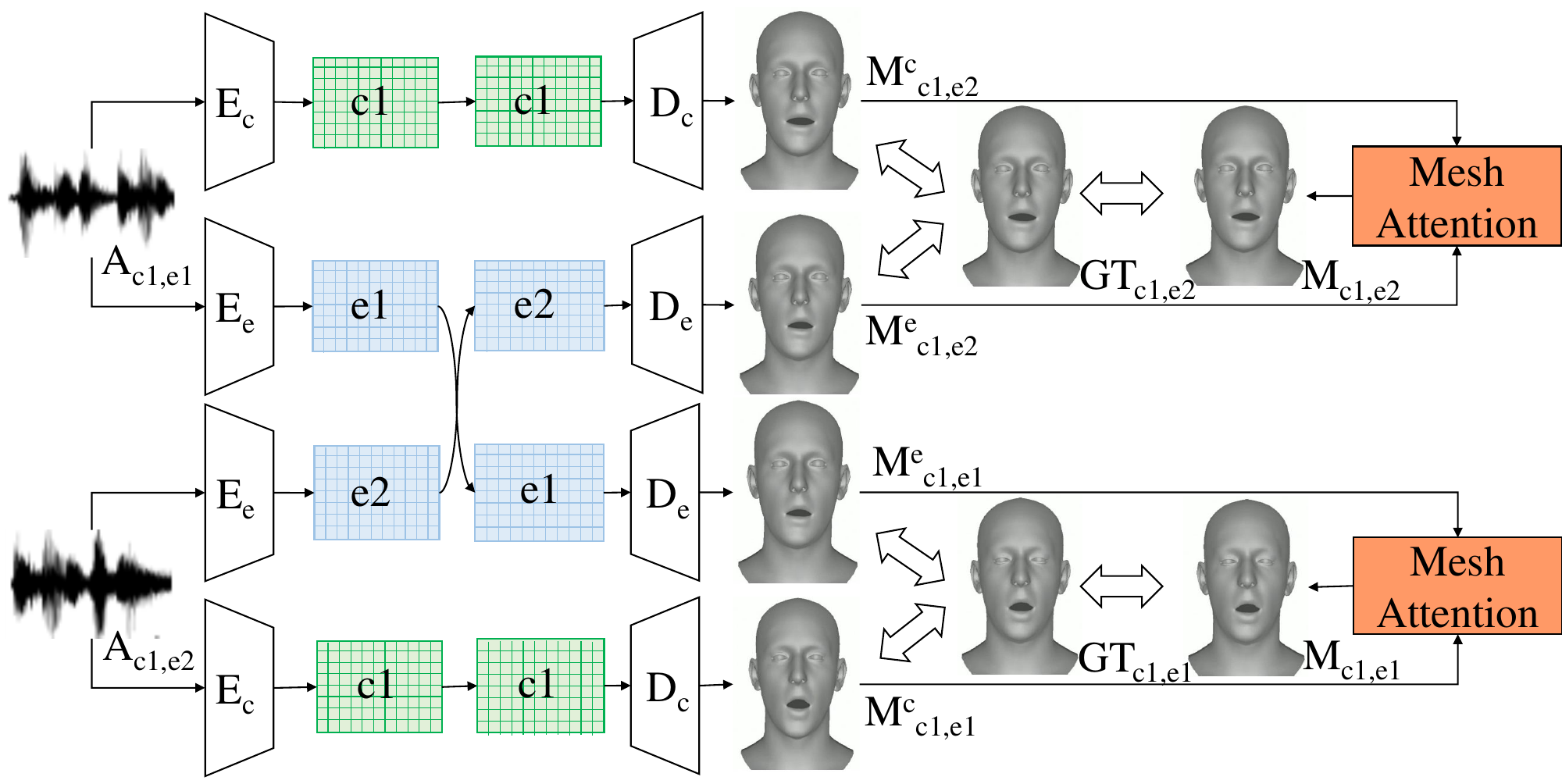}
    \caption{Supervision training strategy of emotion-content disentanglement module. Various inputs of speech, conveying same content and different emotions, are processed to generate cross-reconstructed mesh vertex offsets representing distinct combinations of facial expressions. Supervisions are added to both two branches and the final output.}
    \label{fig5}
\end{figure*}

We set a pseudo-training pair of samples with the same content but different emotions. The emotion features and content features are extracted, respectively, and the two emotion features are exchanged as input of transformer decoder to implement cross reconstruction, as shown in Figure \ref{fig5}. The entire architecture of EmoFace is described in detail as follows: 
\begin{align}\label{eq4}
&F_{c1}=E_{c}(A_{c1,e1}), \\
&F_{e1}=E_{e}(A_{c1,e1}), \\
&F_{e2}=E_{e}(A_{c1,e2}), \\
&M^{c}_{c1,e1}=M^{c}_{c1,e2}=D_{c}(F_{c1}), \\
&M^{e}_{c1,e1}=D_{e}(F_{e1}), M^{e}_{c1,e2}=D_{e}(F_{e2}), \\
&M_{c1,e1}=\epsilon(M^{e}_{c1,e1},M^{c}_{c1,e1}), \\
&M_{c1,e2}=\epsilon(M^{e}_{c1,e2},M^{c}_{c1,e2}),
\end{align}
where $\epsilon$ is the Mesh Attention module, $A_{cx,ey}$ represents audio with content $x$ and emotion $y$, $M_{cx,ey}$ represents the mesh sequence with content $x$ and emotion $y$, $F_c$ and $F_e$ denote the content features and the emotional features, respectively, $E$ and $D$ are the encoder and decoder respectively.

\subsection{Baseline Methods}

\begin{figure*}[t]
    \centering
    \includegraphics[scale=0.5]{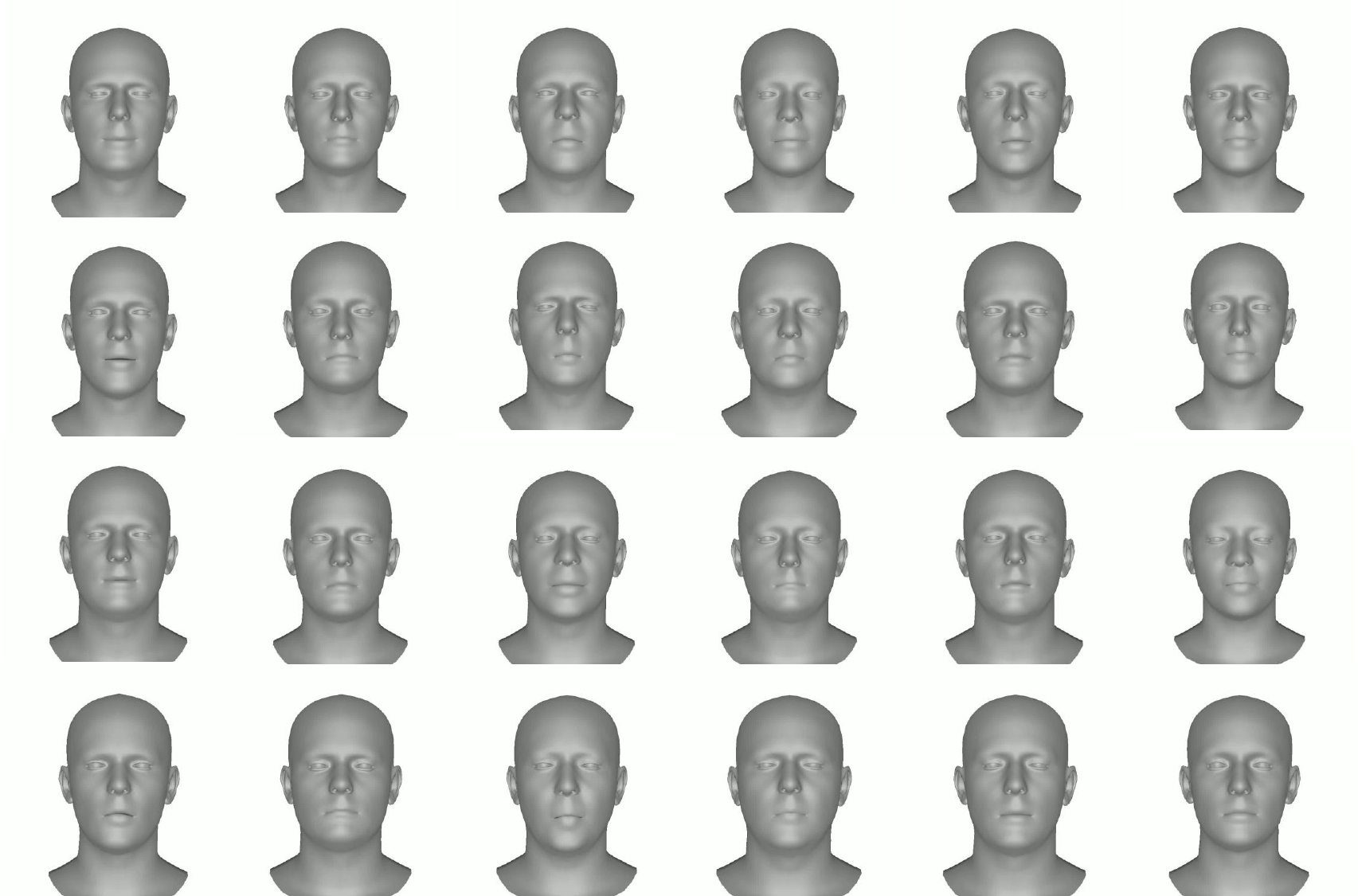}
    \caption{FLAME Head Templates for 24 actors in 3D-RAVDESS. }
    \label{fig6}
\end{figure*}

Our method is compared with FaceFormer \cite{fan2022faceformer}, CodeTalker \cite{xing2023codetalker}, EmoTalk \cite{peng2023emotalk}, SelfTalk \cite{peng2023selftalk} and TalkingStyle \cite{talkingstyle} and achieves state-of-the-art performance on both VOCASET \cite{cudeiro2019capture} and 3D-RAVDESS datasets. Since the official EmoTalk is a blendshape-based method that is incompatible with the marked facial mesh vertices provided by the datasets, we modify its output layer to directly predict the offsets of the vertices. Our method, FaceFormer, CodeTalker and TalkingStyle require conditions on a training speaker identity during inference. Therefore, for unseen subjects in the test dataset, we follow FaceFormer and CodeTalker to obtain the predictions by conditioning on training identities. To adapt the input of non-silent initial frame to the 3D-RAVDESS dataset, we use the first frame as a condition in the training of EmoFace, FaceFormer, CodeTalker and TalkingStyle. All of these models are autoregressive, facilitating the use of input data from the first frame, while SelfTalk and EmoTalk are not autoregressive models. For VOCASET, we use the pre-trained models provided by the five methods to evaluate the vertex error on the VOCA-Test. For 3D-RAVDESS, we retrain other methods under the official experimental configuration. During the evaluation, we compute the error directly between the output vertices and the ground truth. It is worth noting that since CodeTalker training on 3D-RAVDESS crashes no matter how the hyperparameters are adjusted, as mentioned in DF-3DFace \cite{park2023df} and LaughTalk \cite{sung2024laughtalk}. Thus, we follow these two previous work to directly use the pre-trained model trained on VOCASET for zero-shot prediction. 

Besides, The 3D-ETF dataset used in EmoTalk is a blendshape dataset and official EmoTalk has to convert the predict blendshape coefficients into mesh vertices. However, EmoTalk has not yet released its converter. Thus, we modified EmoTalk's output layer to directly predict vertex offsets to adapt to our vertex dataset 3D-RAVDESS to make a fair comparison. 

\subsection{Difference between self-growing and scheduled sampling} 
The primary distinction lies in \textbf{how Stage 2 inputs are constructed}. Scheduled sampling employs a static mixing strategy (Softmax, Gumbel-Softmax, or Sparemax) to blend ground-truth and Stage 1 predicted frames throughout training. In contrast, our self-growing scheme dynamically transitions through three optimized phases (Eq.\ref{4}), each designed to address specific training challenges.

\textbf{Phase 1: Warm-up Period.} The model trains exclusively on groundtruth inputs, avoiding disruptions by unreliable early predictions. This period ensures stable initial learning.

\textbf{Phase 2: Adaptive Prediction Integration.} A cosine annealing scheduler gradually increases the proportion of Stage 1 predictions used as inputs. As prediction accuracy improves with training, the adoption rate accelerates (reflected by the increasing derivative of the cosine function in Eq.\ref{4}), enabling smooth transition to prediction-dominated inputs.

\textbf{Phase 3: Full Prediction Mode.} The system operates solely on Stage 1 predictions, rigorously stress-testing the model against error accumulation – a critical scenario omitted by scheduled sampling.

This phased approach ensures robust training, while scheduled sampling lacks both initial stabilization and final robustness validation phases.

\subsection{Experiments between self-growing and autoregression}
We have compared our method with autoregression in our experiments. After extensive hyperparameter tuning failed to achieve convergence in autoregressive training, we excluded these results from Table \ref{table3}. The non-convergence may be caused by two key factors: (1) the inherent complexity of 3D-RAVDESS dataset samples, which exhibit high variability in emotional expression dynamics, and (2) the error accumulation characteristic of autoregression, where initial prediction inaccuracies amplify catastrophically during sequential generation.

\subsection{3D-RAVDESS Dataset}

In this study, we reconstruct 3D face mesh sequences from 2D videos based on FLAME topology to construct a large 3D emotional talking face dataset, 3D-RAVDESS.

Specifically, 1440 videos from the RAVDESS \cite{livingstone2018ryerson} are processed by converting them into 30 frames per second and capturing the mesh vertices for each frame by EMOCA \cite{danvevcek2022emoca}. To enhance the quality of the dataset and reduce frame-to-frame jitter, a Kalman Filter is applied to the output FLAME parameters, which significantly improves the smoothness of facial animation. The 3D-RAVDESS dataset generates 159,702 frames of mesh vertices, which amounts to approximately 1.5 hours of video content. 

Furthermore, 3D-RAVDESS dataset contains 24 templates for different actors (as shown in Figure. \ref{fig6}). The actors all read sentences to express specific emotions, including neutral, calm, happy, sad, angry, fearful, disgusted, and surprised. A supplementary video demonstrates the animation results of our EmoFace under speech with different emotional types and intensities.

3D-RAVDESS outperforms the dataset used in EMOTE with its strictly controlled multimodal synchronization and standardized actor performances, making it more reliable for precise affective computing research. 

\begin{figure*}[t]
    \centering
    \includegraphics[scale=0.45]{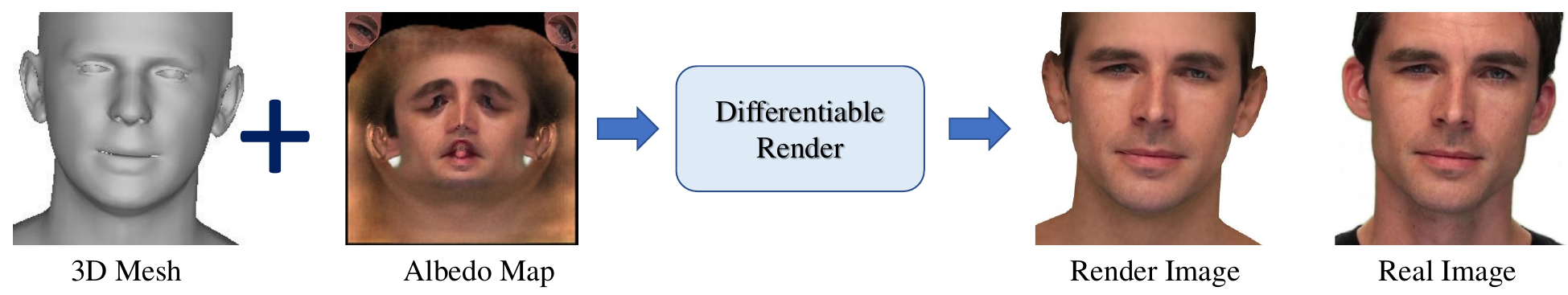}
    \caption{Rendering Process. }
    \label{fig7}
\end{figure*}

\subsection{User study}

We design a comprehensive research questionnaire to evaluate the effectiveness of EmoFace and compare it with FaceFormer, CodeTalker, EmoTalk and SelfTalk. We provide 12 sets of comparison results on 3D-RAVDESS-Test and VOCA-Test with 8 emotions, and finally make 48 questions targeting four aspects: full face realism, lip synchronization, eye movement and emotion expression. The questionnaire shows comparison videos to the respondents and asks them to rate the effectiveness of each algorithm. We calculate the Mean Opinion Score (MOS) of all methods and EmoFace obtains the highest MOS, as shown in \ref{table5}, suggesting that our method receives the most positive feedback. 

\begin{table}[t]

    \centering
        \scalebox{0.75}{
        \begin{tabular}{@{}lcccc@{}}
        \toprule
        Methods     & \begin{tabular}[c]{@{}c@{}}full face \\ realism\end{tabular} & \begin{tabular}[c]{@{}c@{}}lip  \\ synchronization\end{tabular} & \begin{tabular}[c]{@{}c@{}}eye\\ movement\end{tabular}  &\begin{tabular}[c]{@{}c@{}}emotion\\ expression\end{tabular}  \\ \midrule
        FaceFormer    & 3.3    & 3.3  &  3.0     & 3.2 \\
        CodeTalker    & 2.6    & 2.4 &  2.3     & 2.4 \\
        EmoTalk      &  3.5   & 3.2 & 3.6       & 3.9 \\
        SelfTalk      & 3.7    & 3.6 &  3.6      & 3.8 \\       
        \textbf{EmoFace}    & \textbf{4.0}   & \textbf{3.9}  & \textbf{3.9}  & \textbf{3.9}\\ 
        \hline

        \end{tabular}
}     
    \caption{User study results.}
    \label{table5}
\end{table}

\subsection{Application}
The final render results with textures can be obtained based on the output mesh of our method. The rendering process of transforming a 3D mesh into a final rendered image is shown in Figure. \ref{fig7}. The requirements include materials, textures and scene setup, each contributing to the visual fidelity and realism of the final output.

\subsubsection{Materials and Textures}
Materials and textures are applied to the mesh to define its visual properties. On the one hand, materials define surface characteristics such as color, reflectivity and transparency, which determine how the mesh interacts with light. On the other hand, textures add details like patterns and surface irregularities. In our work, we use AlbedoGAN \cite{rai2024towards} to obtain an albedo map image, a texture image that represents the color information without any lighting effects.

\subsubsection{Scene Setup}
The scene setup involves placing the mesh within a virtual environment along with cameras, lighting and shading. Camera setup defines the camera's position, orientation, and properties to capture the scene appropriately. And lighting adds light sources to illuminate the scene, which significantly impacts the final appearance of the mesh. The shading calculates the color and intensity of each pixel based on lighting and material properties. 

\subsubsection{Rendering}
A differentiable renderer is applied for high-quality renderings from the mesh and the albedo map.

\subsection{Supplementary Video}

We provide a Supplementary Video showing the results under VOCASET-Test, 3D-RAVDESS-Test and long speech in the wild, performance on different emotions and comparative demonstrations of ablation experiments.

\end{document}